\documentclass[a4paper,conference]{IEEEtran}
\IEEEoverridecommandlockouts

\usepackage{cite}
\usepackage{amsmath,amssymb,amsfonts}
\usepackage{graphicx}
\usepackage{textcomp}
\usepackage{xcolor}
\usepackage{multirow}%
\usepackage{amsthm}%
\usepackage{mathrsfs}%
\usepackage{textcomp}%
\usepackage{manyfoot}%
\usepackage{booktabs}%
\usepackage{algorithm}%
\usepackage{algorithmicx}%
\usepackage{algpseudocode}%
\usepackage{listings}%
\usepackage{epsfig} % for postscript graphics files
\usepackage{mathptmx} % assumes new font selection scheme installed
\usepackage{times} % assumes new font selection scheme installed
\usepackage{textcomp}
\usepackage{gensymb}
\usepackage{caption}
\usepackage{hyperref}
\usepackage{mwe}
\usepackage{verbatim}
\usepackage{amsbsy}
\usepackage{siunitx}
\usepackage{tabularx,booktabs}
\usepackage{dblfloatfix}
\usepackage{bm}
\usepackage{subcaption}
\usepackage{float}
\usepackage{microtype}

\usepackage{epstopdf} % Converte imagens EPS para PDF para compressão
\def\BibTeX{{\rm B\kern-.05em{\sc i\kern-.025em b}\kern-.08em
    T\kern-.1667em\lower.7ex\hbox{E}\kern-.125emX}}
\begin{document}

\title{From Seedling to Harvest: The GrowingSoy Dataset\\for Weed Detection in Soy Crops\\via Instance Segmentation}

\author{
\IEEEauthorblockN{1\textsuperscript{st} Raul Steinmetz}
\IEEEauthorblockA{\textit{Universidade Federal de Santa Maria}\\
Santa Maria, Brazil \\
rsteinmetz@inf.ufsm.br}
\and
\IEEEauthorblockN{2\textsuperscript{nd} Victor Augusto Kich}
\IEEEauthorblockA{\textit{University of Tsukuba}\\
Tsukuba, Japan \\
victorkich98@gmail.com}
\and
\IEEEauthorblockN{3\textsuperscript{rd} Henrique Krever}
\IEEEauthorblockA{\textit{Universidade Federal de Santa Maria}\\
Santa Maria, Brazil \\
hlkrever@inf.ufsm.br}
\and
\IEEEauthorblockN{4\textsuperscript{rd} João Davi Rigo Mazzarolo}
\IEEEauthorblockA{\textit{Universidade Federal de Santa Maria}\\
Santa Maria, Brazil \\
jdmazzarolo@inf.ufsm.br}
\and
\IEEEauthorblockN{5\textsuperscript{th} Ricardo Bedin Grando}
\IEEEauthorblockA{\textit{Universidad Tecnológica del Uruguay}\\
Rivera, Uruguay \\
ricardo.bedin@utec.edu.uy}
\and
\IEEEauthorblockN{6\textsuperscript{th} Vinicius Marini}
\IEEEauthorblockA{\textit{Universidade Federal de Santa Maria}\\
Santa Maria, Brazil \\
vinicius.marini@ufsm.br}
\and
\IEEEauthorblockN{7\textsuperscript{th} Celio Trois}
\IEEEauthorblockA{\textit{Universidade Federal de Santa Maria}\\
Santa Maria, Brazil \\
trois@inf.ufsm.br}
\and 
\IEEEauthorblockN{8\textsuperscript{th} Ard Nieuwenhuizen}
\IEEEauthorblockA{\textit{Wageningen University \& Research}\\
Wageningen, Netherlands \\
ard.nieuwenhuizen@wur.nl}
}

\maketitle

\begin{abstract}
Deep learning, particularly Convolutional Neural Networks (CNNs), has gained significant attention for its effectiveness in computer vision, especially in agricultural tasks. Recent advancements in instance segmentation have improved image classification accuracy. In this work, we introduce a comprehensive dataset for training neural networks to detect weeds and soy plants through instance segmentation. Our dataset covers various stages of soy growth, offering a chronological perspective on weed invasion's impact, with 1,000 annotated images. To validate our data, we also provide 6 state of the art models, trained in this dataset, that can understand and detect soy and weed in every stage of the plantation process, the best results achieved were a segmentation average precision of 79.1\% and an average recall of 73.3\% across all plant classes. Moreover, the YOLOv8M model attained 78.7\% mean average precision (mAp-50) in caruru weed segmentation, 69.6\% in grassy weed segmentation, and 90.1\% in soy plant segmentation.\footnote{Video of the experiments is available at: https://youtu.be/Ir0vwtAk8Tc}\footnote{GitHub Repository is available at: https://github.com/raulsteinmetz/soy-segmentation}\footnote{SoyGrowing dataset is available at: https://github.com/raulsteinmetz/soy-segmentation-ds}
\end{abstract}


\section{Introduction}\label{sec:introduction}

\begin{figure*}[tbp]
    \centering
    \includegraphics[width=\linewidth]{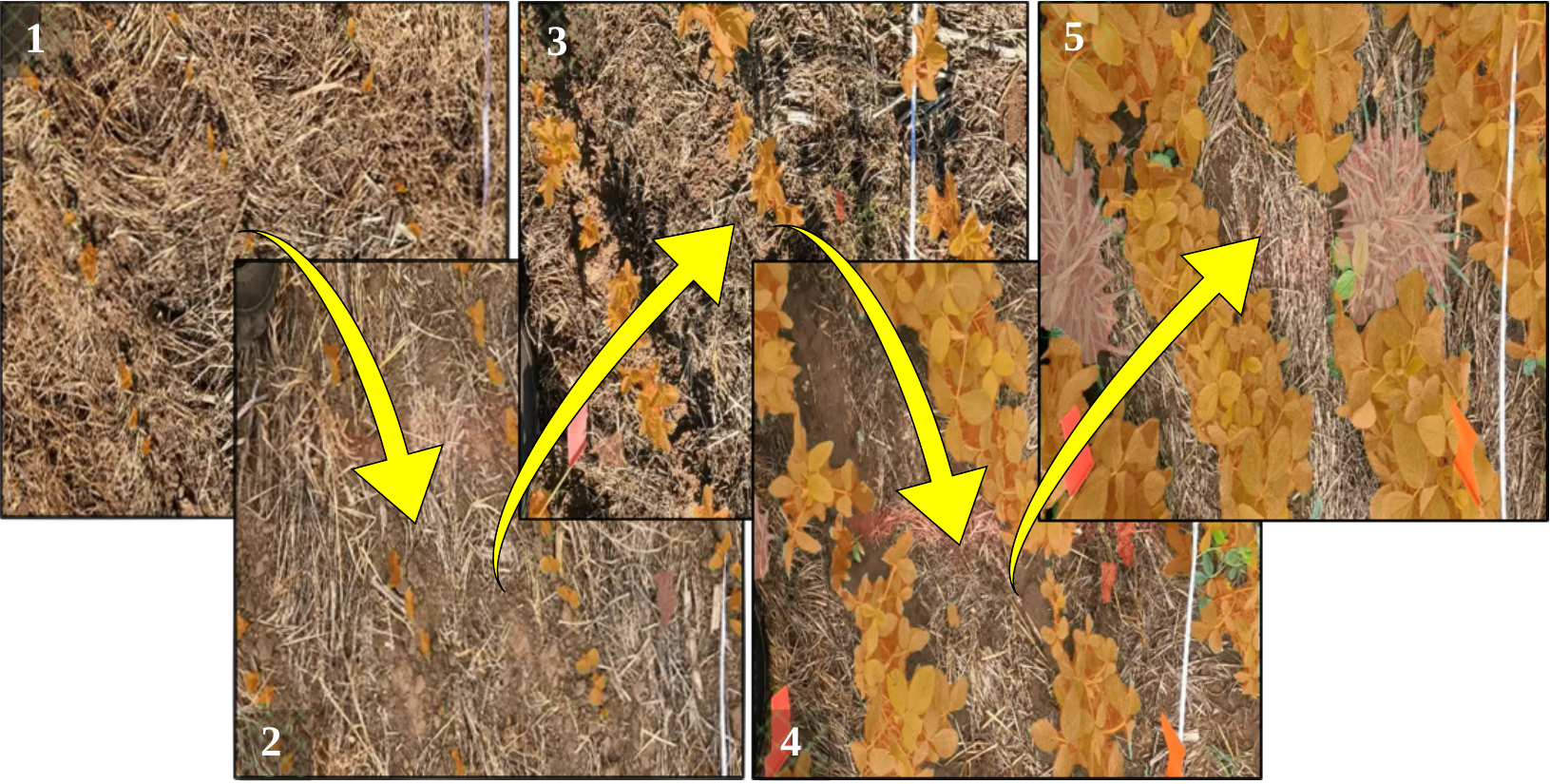}
    \caption{Some examples of instance segmentation on the GrowingSoy dataset. Arrows highlight the transitions between temporal states, encompassing all growth stages of the soy plant.}
    \label{fig:temporal}
\end{figure*}

Effective crop management is essential for maximizing yields, particularly in environments susceptible to issues like weed infestations, such as soybean crops. Despite efforts to control weeds through methods such as chemical applications, the persistent threat of unexpected weed invasions remains a significant risk to crop quality and yield. Therefore, vigilant and responsive agricultural practices are crucial.

Soybeans occupy a pivotal role globally, being the second most produced vegetable oil and serving as a major source of protein for both animal feed and human consumption~\cite{pagano2016importance}. Effective management techniques that enable early identification and removal of weeds are essential for increasing crop production.

In recent years, advancements in hardware and software technologies have propelled neural networks into the spotlight as powerful tools capable of automating crucial agricultural processes~\cite{kujawa2021artificial,escamilla2020applications}. These innovations include autonomous robot control~\cite{williams2019robotic,vulpi2021recurrent} and computer vision algorithms, such as Convolutional Neural Networks (CNNs) for image classification~\cite{rawat2017deep}. These technologies facilitate automated disaster detection while simultaneously conducting a comprehensive analysis of plantation status and key metrics~\cite{kamilaris2018review}.

A recent advancement in computer vision is instance semantic segmentation~\cite{long2015fully,guo2018review}, a method that divides images into distinct segments and entities with pixel-level accuracy, enhancing object recognition and delineation. This technique has significant potential in agricultural computer vision, where accurate identification and analysis of individual components within a crop field are essential for efficient crop management. However, the scarcity of properly labeled and processed data for segmentation remains a challenge for the development of advanced applied solutions.

The prevailing issue in crop datasets lies in the lack of segmentation labels, as the majority remain unlabeled for this purpose. Notably, datasets such as DeepWeeds~\cite{olsen2019deepweeds}, Weed25~\cite{wang2022weed25}, and the Soybean images dataset ~\cite{MIGNONI2022107756} are exclusively annotated for image classification, overlooking the segmentation aspect. Furthermore, a significant portion of these datasets consists of images with low resolution and inadequate illumination.

% In contrast, we introduce a dataset that addresses these limitations by providing a substantial number of high-quality images depicting soy and weeds. What sets our dataset apart is the inclusion of segmentation labels for all images. This not only contributes to a more comprehensive dataset but also fills a crucial gap in the availability of well-annotated crop datasets for segmentation tasks. A set of segmented images using a model trained on our dataset are depicted in Fig.~\ref{fig:temporal}.
In contrast, we introduce a dataset that addresses these limitations by providing a 1000 high-quality images depicting soy and weeds. What distinguishes our dataset is the inclusion of segmentation labels for all images. This enhancement contributes to a more comprehensive dataset and fills a crucial gap in the availability of well-annotated crop datasets for segmentation tasks. A set of segmented images using a model trained on our dataset is depicted in Fig.~\ref{fig:temporal}.

%Hence, this research aims to contribute with the context presented through the following contributions. We investigate two leading frameworks within the domain of computer vision segmentation YOLOv5 and YOLOv8. Our study encompasses a variety of model sizes, including medium, large, and x-large, to serve as the foundation of our approach to instance segmentation. These frameworks, underpinned by cutting-edge research and technological advancements, have played a pivotal role in shaping the landscape of real-time object detection and segmentation methodologies. Our exploration goes beyond the surface, aiming to offer a deeper understanding of their architectural intricacies and operational modalities, recognizing their profound impact on detecting both soybean and weed.

Hence, the contributions of this work are:

\begin{itemize}
    \item We present a novel dataset with 1,000 annotated and high-resolution images for training instance segmentation neural networks, focusing on weed-infested soy plantations. The dataset spans the entire soy growth process, from initial stages to maturity with pervasive weed presence.
    
    \item Additionally, to validate the dataset, we provide six state-of-the-art neural network models and perform an in-depth comparison among them, leveraging this temporal dataset to achieve impressive mean average precision in weed and soy segmentation across all soy plantation stages.
\end{itemize}

The structure of this paper is as follows: Section~\ref{sec:related_works} offers a literature review, delving into previous research in computer vision for crop monitoring and instance segmentation. In Section~\ref{sec:dataset}, we outline our methodology for creating the dataset. In Section~\ref{sec:neural_nets}, we depict the process of selecting and training the neural networks involved in the study. Section~\ref{sec:results} presents the results obtained using the neural networks on our dataset. Section~\ref{sec:analysis} presents an analysis of the performance in images that contain both soy and weed, in seedling and harvest plantation stages. Section~\ref{discussion} present some clarifications about the results and analysis. Finally, Section~\ref{sec:conclusion} summarizes our conclusions drawn from this study, highlighting the significance of our findings for agricultural management and the potential for further advancements in instance segmentation for crop monitoring.

\section{Related Work}\label{sec:related_works}

Over the years, continuous advancements in artificial intelligence have enhanced the widespread use of CNNs in precision agriculture and the use of instance segmentation for computer vision. In this session, we will explore some works that obtained success employing this technology.

In the field of plant disease detection, leveraging the use of a deep CNN, Francis and Deisy~\cite{francis2019disease} attained an 87\% precision rate in the detection and classification of diseases in apple and tomato leaf images. They proposed a methodology consistent of an input image of 64x64 pixels and a smaller amount of parameters when compared with the other models that can be used for this problem. Similarly, Bedi and Gole~\cite{bedi2021plant} developed a hybrid AI model that combines a convolutional autoencoder and a CNN to identify the presence of Bacterial Spot disease in peach plants by analyzing images of their leaves, the model avoids immense quantities of parameters and also increases overall accuracy. Their main contribution was related to the small amount of parameteres (less then 10K parameters) reducing the amount of time needed for training and to perform the real time detection of possible plant deceases.

Weed detection through CNN's has been widely adopted in the literature~\cite{hasan2021survey}. Ramirez \textit{et al.}~\cite{ramirez2020deep} employed the DeepLabv3 neural network architecture to identify weed presence within a beet agricultural field by analyzing high-resolution aerial imagery. For the latter, a comparison between three distinct model was done, concluding that the DeepLabv3 presented a better performance, besides being the one that needs the most computational resource. In a similar vein, Hu \textit{et al.}~\cite{hu2021rice} utilized the YOLOv4 architecture to identify 12 distinct types of weeds in a rice plantation. A comparative analysis with the YOLOv3 architecture was conducted, demonstrating enhancements in multiple metrics.

Several datasets containing images of both weeds and crops have been made available in the literature like the one presented by Sudars \textit{et al.}~\cite{sudars2020dataset}, which consists of 1118 annotated images encompassing six different crop types and eight different weed types. 
Another notable dataset is the TobSet, presented by Alam \textit{et al.}~\cite{alam2022tobset}. It is a collection of images featuring tobacco plants and weeds captured from local fields, encompassing diverse growth stages and varying lighting conditions. The dataset includes 7000 images of tobacco plants and 1,000 images of weeds.

In the domain of crop monitoring, another significant application of CNNs is in the detection of plant diseases. Medhi and Deb~\cite{medhi2022psfd} introduced an image dataset featuring various banana plant varieties and the diseases associated with them, summing up to an extensive collection of over 8000 images for deep neural network training. In a similar vein, Moupojou \textit{et al.}~\cite{moupojou2023fieldplant} introduced the FieldPlant dataset, consisting of 5,170 plant disease images acquired directly from plantations. These images were annotated at the leaf level by plant pathologists. The study demonstrates that FieldPlant surpasses current datasets in terms of performance in classification tasks, establishing itself as a more precise tool for disease detection in practical agricultural scenarios.

Leveraging the power of semantic segmentation, various computer vision tasks can have several improvements in accuracy. Yang \textit{et al.}~\cite{yang2023semantic} conducted a comprehensive study comparing instance segmentation methods on established datasets, alongside an examination of practical applications of semantic segmentation in autonomous driving. Also using instance segmentation concepts, Zhou \textit{et al.}~\cite{zhou2023hybrid} introduced an innovative approach, a combination of Swin Transformer and CNN, for tunnel lining crack detection using instance segmentation. This method achieved a mean intersection over union (mIoU) of 77.41\% and a mean pixel accuracy of 84.42\% on the custom datasets presented in this paper. Notably, these results outperform previous CNN-based and transformer-based semantic segmentation algorithms in terms of segmentation accuracy.

Semantic segmentation shows notable enhancements in various fields of crop management and monitoring. Anand \textit{et al.}~\cite{anand2021agrisegnet} introduced AgriSegNet, a deep learning framework designed for the automatic identification of farmland anomalies, including standing water and weed clusters, by employing multiscale attention-based semantic segmentation on UAV-captured images. This framework is tailored for IoT-assisted precision agriculture. Bosilj \textit{et al.}~\cite{bosilj2020transfer} conducted a study examining the feasibility of transfer learning across different crop types, demonstrating the potential to reduce training time by up to 80\% in semantic segmentation-based architectures.

To the best of our knowledge, no dataset in the literature provides RGB, high-quality images of soy plants along with human-annotated semantic segmentation labels, covering their growth stages from seedling to harvest. As well as encompassing the presence of weed invasions such as caruru and grassy weeds. The temporal factor in the dataset is especially important since you can track the plantation as well as the weeds invasion, for a better course of action regarding ways to deal with the weeds. Furthermore, this paper represents the pioneering effort to employ state-of-the-art YOLO architectures for accurate segmentation of both soy and weeds throughout all growth stages of these plants.

\section{The Dataset}\label{sec:dataset}

Creating the soy instance segmentation dataset included recording videos, selecting images, and manually labeling for instance segmentation. This section illustrates these processes.

\subsection{Data Gathering}

\begin{figure}[b!]
    \vspace{-5mm}
    \centering
    \includegraphics[width=\linewidth]{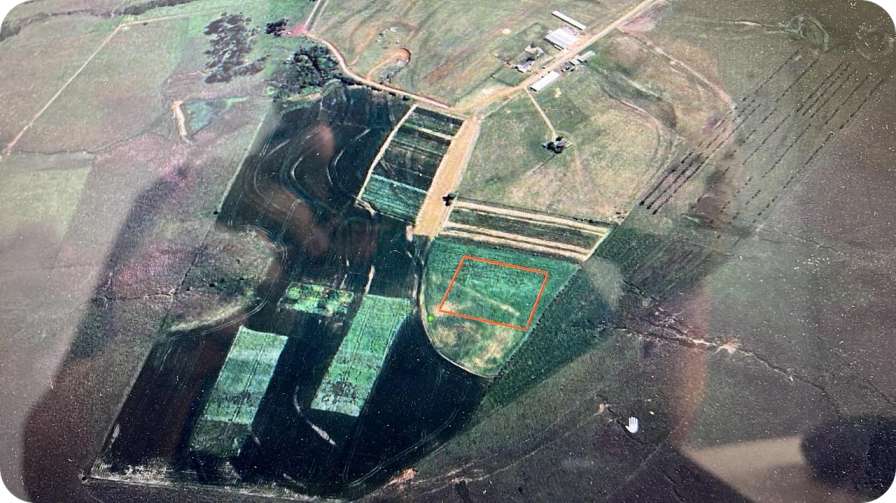}
    \caption{Location where the videos were recorded.}
    \label{fig:location}
\end{figure}

The data collection process involved recording multiple videos that captured various distinct stages of the soy crop development, spanning from the initial growth phases to more advanced stages. These videos also documented the presence of caruru and grassy weeds. The selected areas of the plantation where the videos were shot received different levels of chemical treatment, including both medium and chemical-free approaches. This recording took place at the Universidade Federal de Santa Maria, within a dedicated soy plantation for research, located at approximately 29° 43' 42'' S, 53° 45' 24'' W, an aerial image can be seen in Fig.~\ref{fig:location}.

\begin{figure}[t!]
    \centering
    \includegraphics[width=\linewidth]{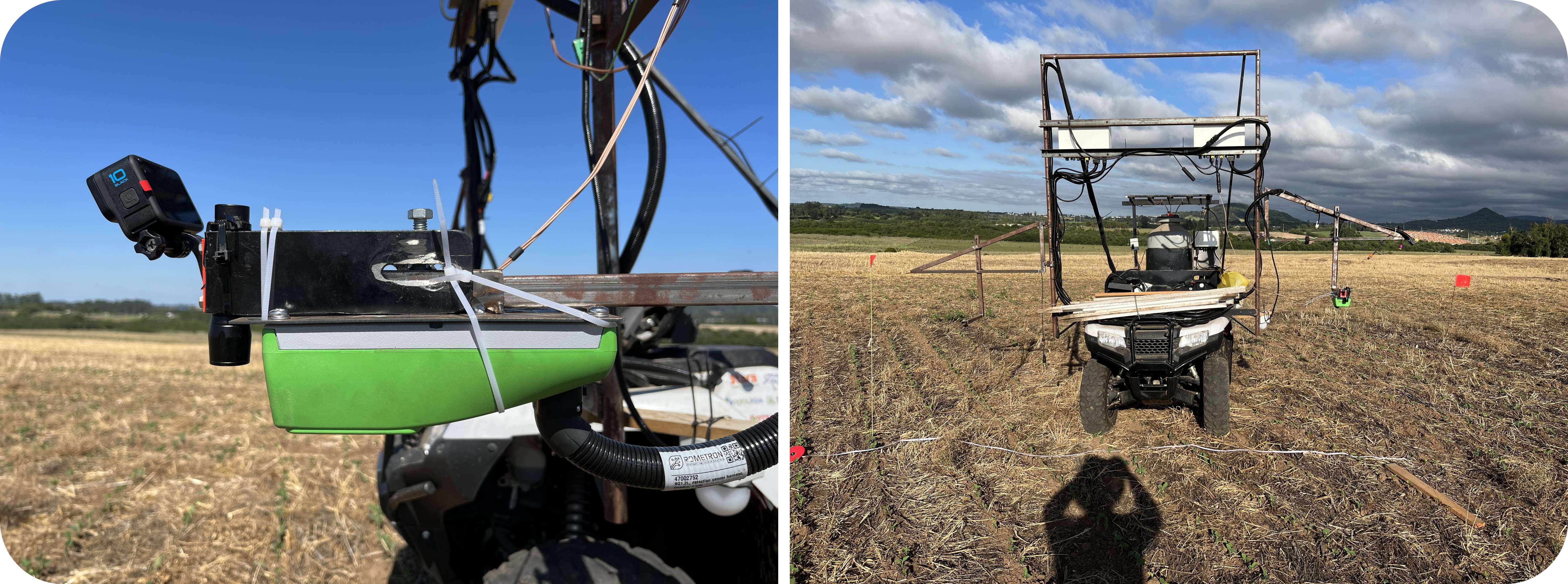}
    \caption{Setup used for the video recordings.}
    \label{fig:setup}
\end{figure}

The videos were captured using a GoPro Hero 12 camera with 4K resolution and a frame rate of 120 frames per second. To ensure comprehensive coverage, the camera was mounted on an ATV four-wheeled vehicle, which followed a consistent 14-meter path through the plantation at a steady pace of 2 kilometers per hour, recording the crop's growth at various intervals. The ATV and camera setup can be seen in Fig.~\ref{fig:setup}.

\subsection{Dataset Creation and Annotation}

To curate a representative database of soy plantation stages, we selected 1000 frames from recorded videos. A sequential extraction, with a fixed step across all videos, ensured comprehensive coverage of each stage in the soy plantation life cycle. Through iterative experimentation, we determined the optimal number of images for model training, ranging from 250 to 1500 images. Interestingly, discernible model improvement plateaued after reaching 1000 images, affirming that this quantity sufficiently encapsulates the task's complexity.

\begin{figure}[b!]
    \centering
    \includegraphics[width=\linewidth]{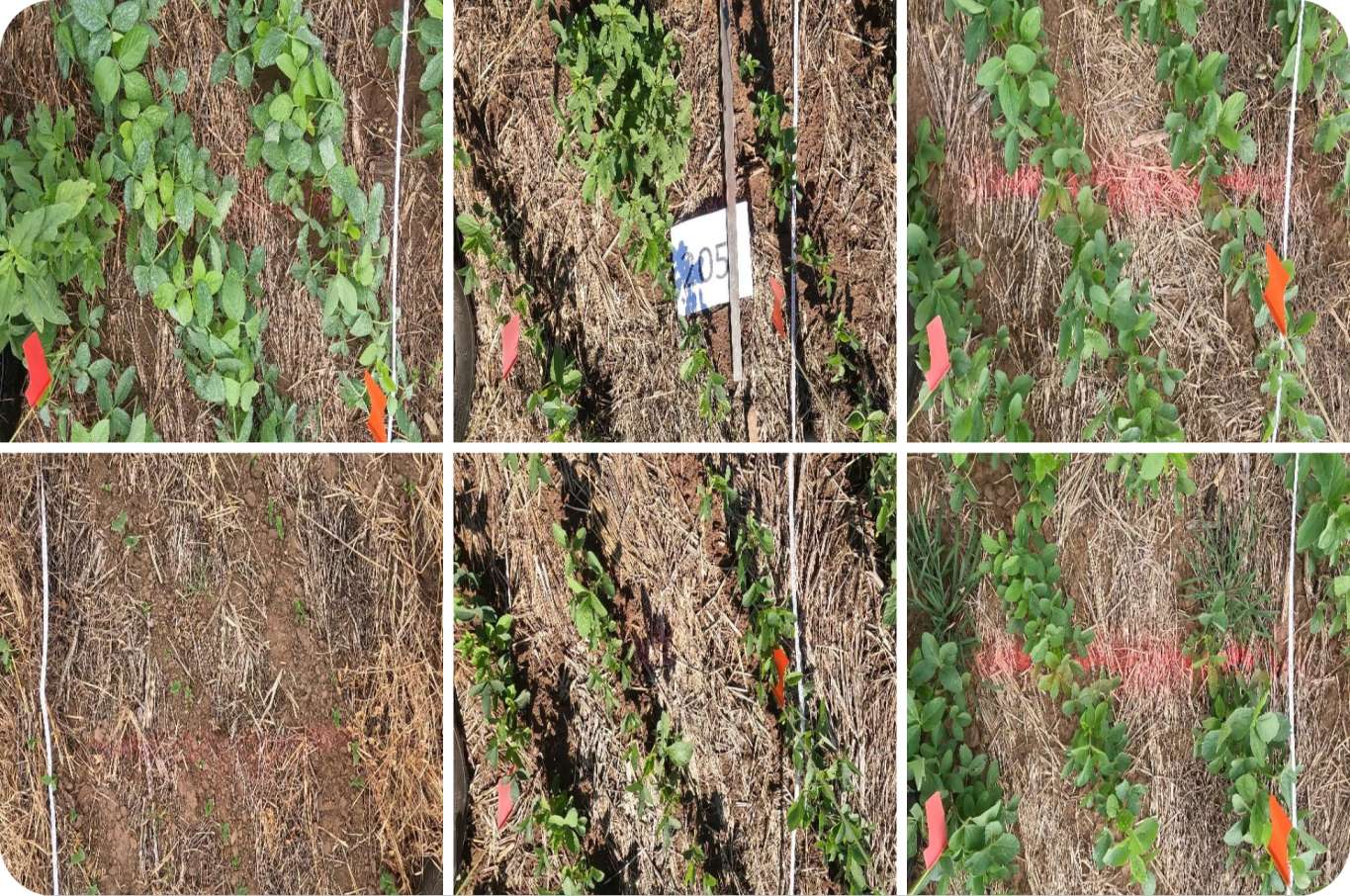}
    \caption{Image sample from our dataset. Different stages of the plantation can be easily visualized.}
    \label{fig:image-sample}
\end{figure}

Subsequently, the selected frames, initially with 4992 x 2496 pixels, underwent resizing to 640x640 pixels using cubic interpolation, aligning them with the specifications required for neural network models. To provide a glimpse into the dataset, a subset of these images is showcased in Fig.~\ref{fig:image-sample}.

In the labeling process, instance segmentation was employed. This technique involved delineating and fitting polygons around all the objects, which in this case were the plants within the scenes. This detailed labeling approach allowed the neural networks to learn how to accurately segment each individual plant, enabling precise object recognition and classification within the dataset. The labeling was performed manually using the Roboflow Framework~\cite{roboflow2023}, which streamlines the creation of such labeled datasets for machine learning purposes \cite{lin2022roboflow}.

\section{Models Trained for Data Validation} \label{sec:neural_nets}

In this section, we analyze two significant frameworks in image segmentation, used to validate the quality of our data: YOLOv5~\cite{yolov52020} and YOLOv8~\cite{ultralytics2023}. These architectures and their various size models (medium, large, and x-large) were chosen for our instance segmentation approach.

YOLOv5, developed by Ultralytics, is recognized for its real-time object detection capabilities. It divides an image into a grid and predicts bounding boxes and class probabilities for each grid cell using anchor boxes, which are predefined shapes for predicting object shapes. YOLOv5's success is largely due to its flexible Pythonic structure, enabling rapid community-driven improvements.

YOLOv8, the latest model in the YOLO series, also developed by Ultralytics, supports object detection, image classification, and instance segmentation. It introduces architectural enhancements over YOLOv5, including anchor-free detection that predicts object centers directly, reducing the number of box predictions and speeding up the Non-Maximum Suppression (NMS) process. YOLOv8 also features new convolutions and the C2f module, and it incorporates online augmentations during training, such as mosaic augmentation.

The primary distinction between YOLOv5 and YOLOv8 lies in their approach to object detection. YOLOv5 uses anchor boxes, while YOLOv8 employs anchor-free detection, predicting object centers directly. This approach addresses the limitations of anchor boxes, which may not represent the distribution of custom datasets. YOLOv8 also introduces new convolutions and modules, such as the C2f module, and utilizes online augmentations like mosaic augmentation, enhancing the model's ability to recognize objects under various conditions. In terms of accuracy, YOLOv8 has shown improvements over YOLOv5, particularly on benchmarks like COCO and Roboflow 100.

\section{Evaluation and Results}\label{sec:results}

\begin{figure*}[t!]
     \centering
\begin{subfigure}{0.321\textwidth}
    \includegraphics[width=\linewidth]{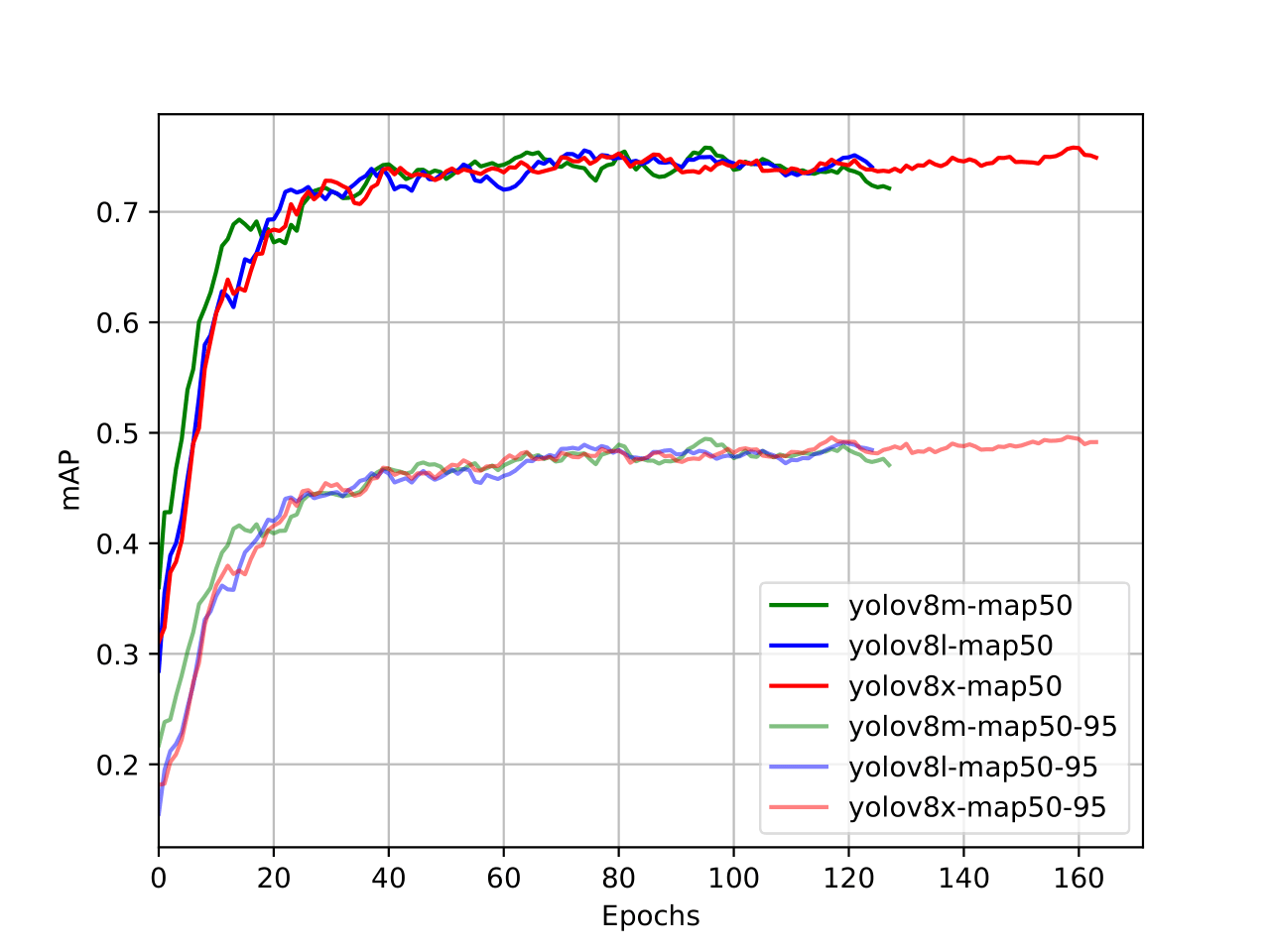}
    \caption{YOLOv8 mAP increase.}
    \label{fig:treining-1}
\end{subfigure}
~
\begin{subfigure}[b]{0.321\textwidth}
    \includegraphics[width=\linewidth]{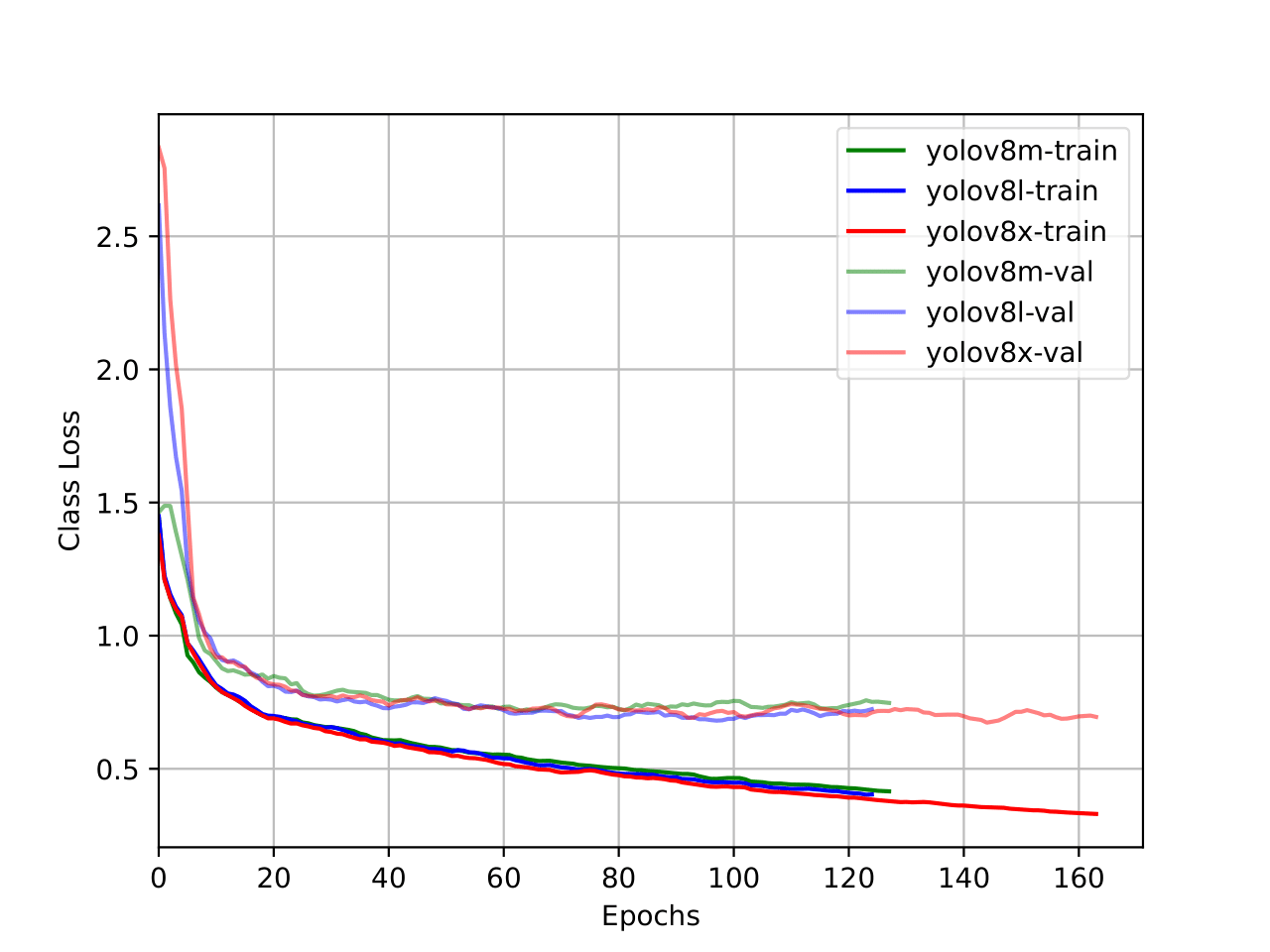}
    \caption{YOLOv8 class loss decrease.}
    \label{fig:treining-2}
\end{subfigure}
~
\begin{subfigure}[b]{0.321\textwidth}
    \includegraphics[width=\linewidth]{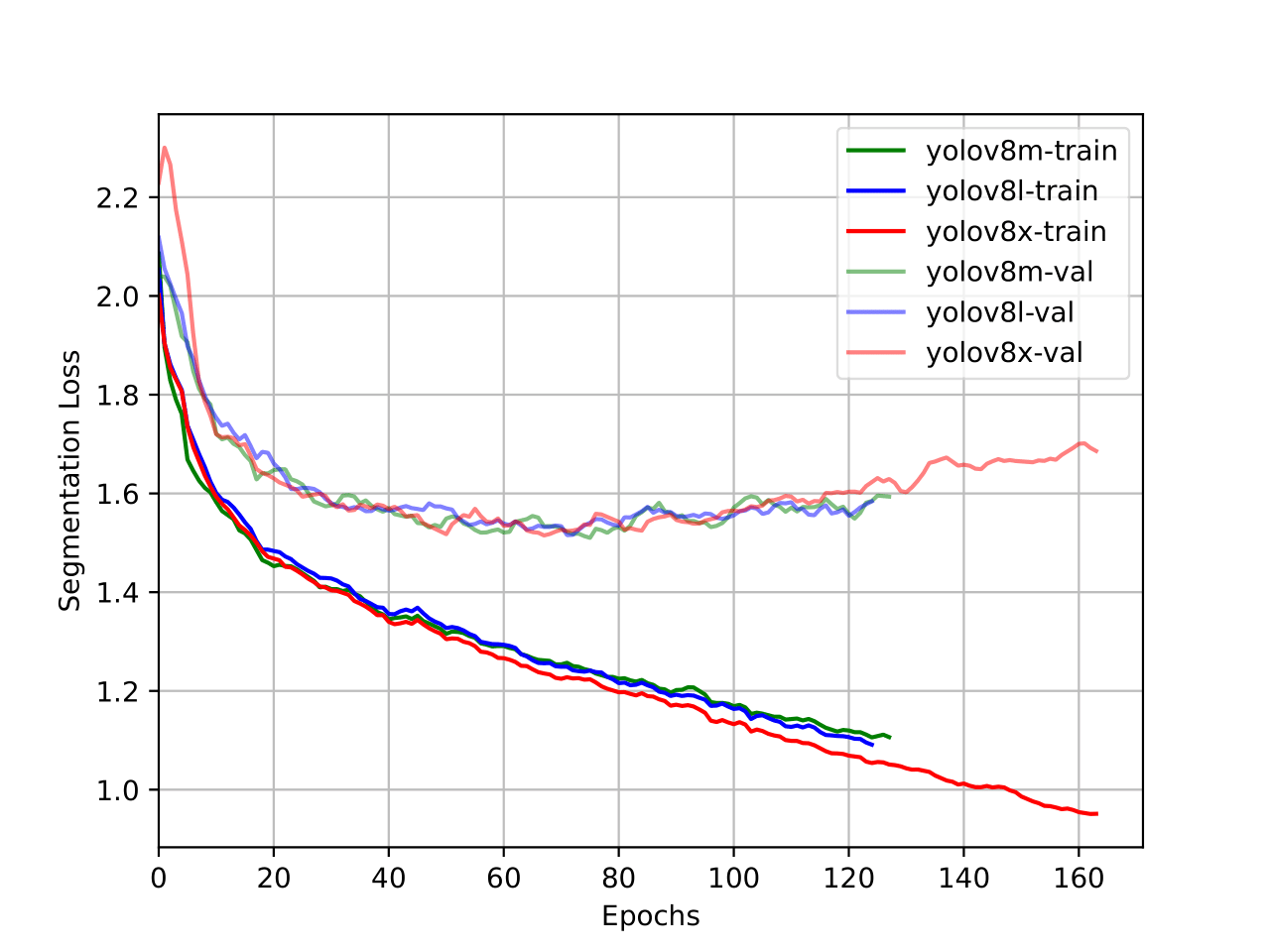}
    \caption{YOLOv8 segmentation loss decrease.}
    \label{fig:treining-3}
\end{subfigure}
\\
\begin{subfigure}[b]{0.321\textwidth}
    \includegraphics[width=\linewidth]{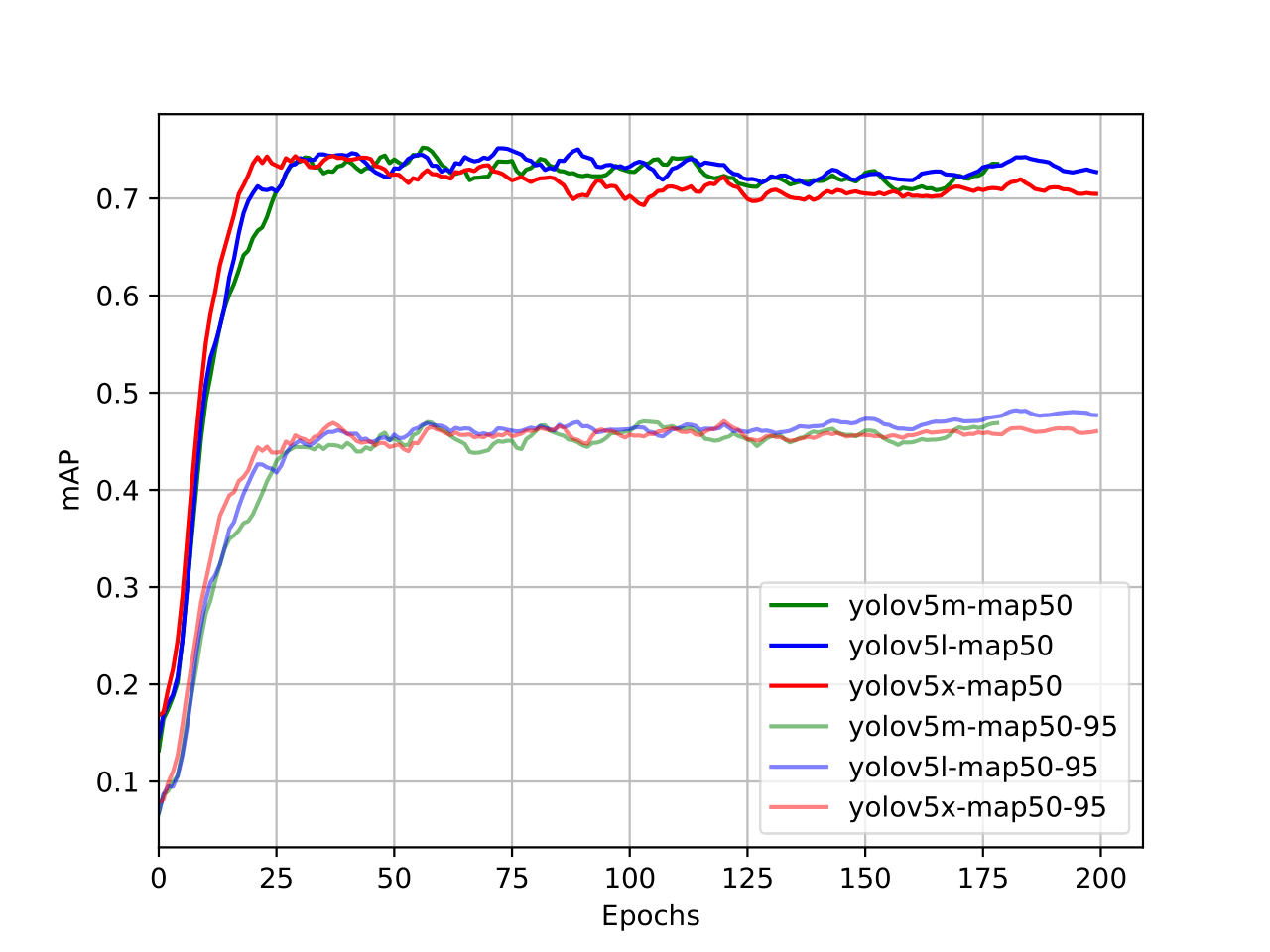}
    \caption{YOLOv5 mAP increase.}
    \label{fig:treining-4}
\end{subfigure}
~
\begin{subfigure}[b]{0.321\textwidth}
    \includegraphics[width=\linewidth]{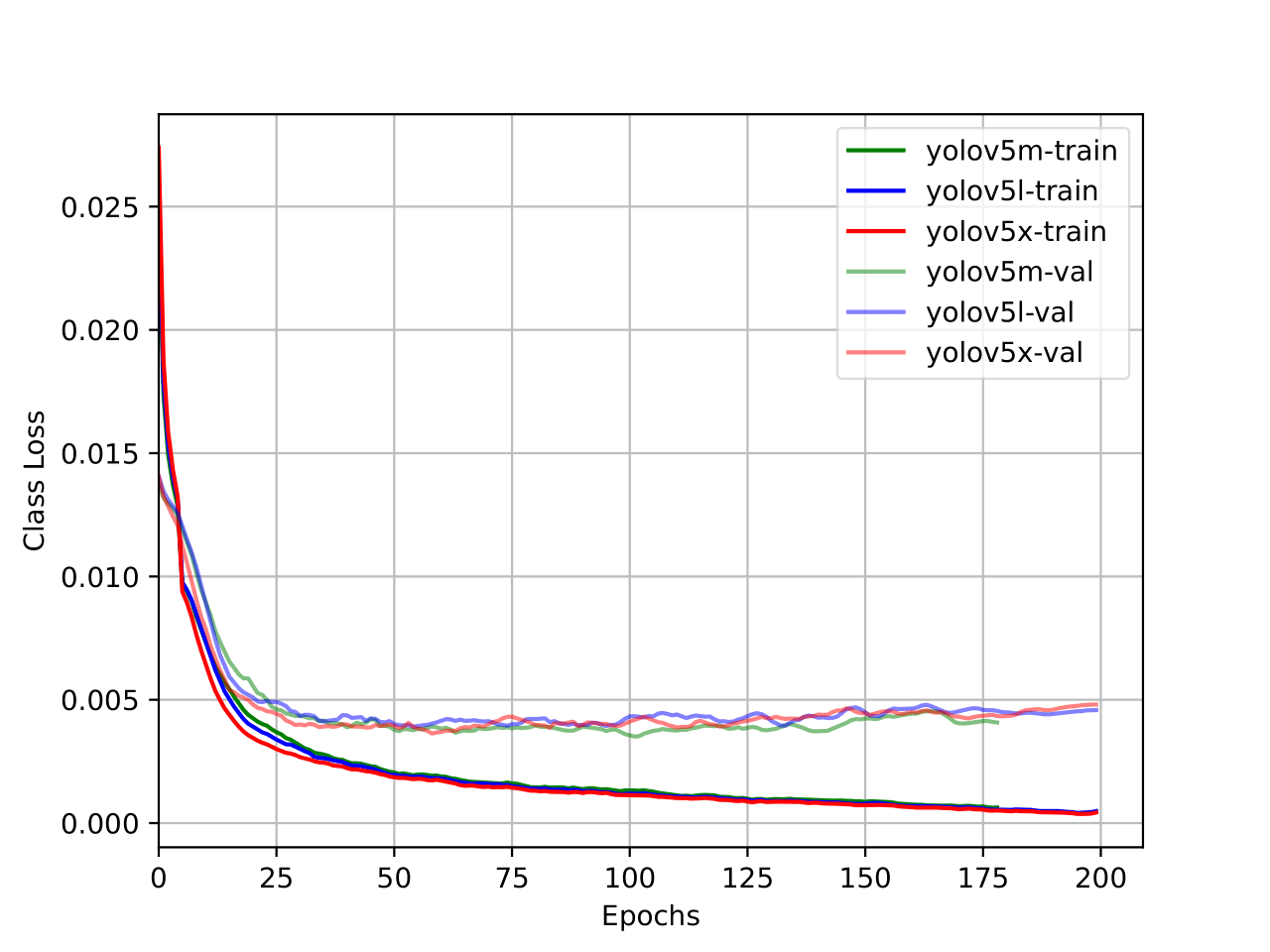}
    \caption{YOLOv5 class loss decrease.}
    \label{fig:treining-5}
\end{subfigure}
~
\begin{subfigure}[b]{0.321\textwidth}
    \includegraphics[width=\linewidth]{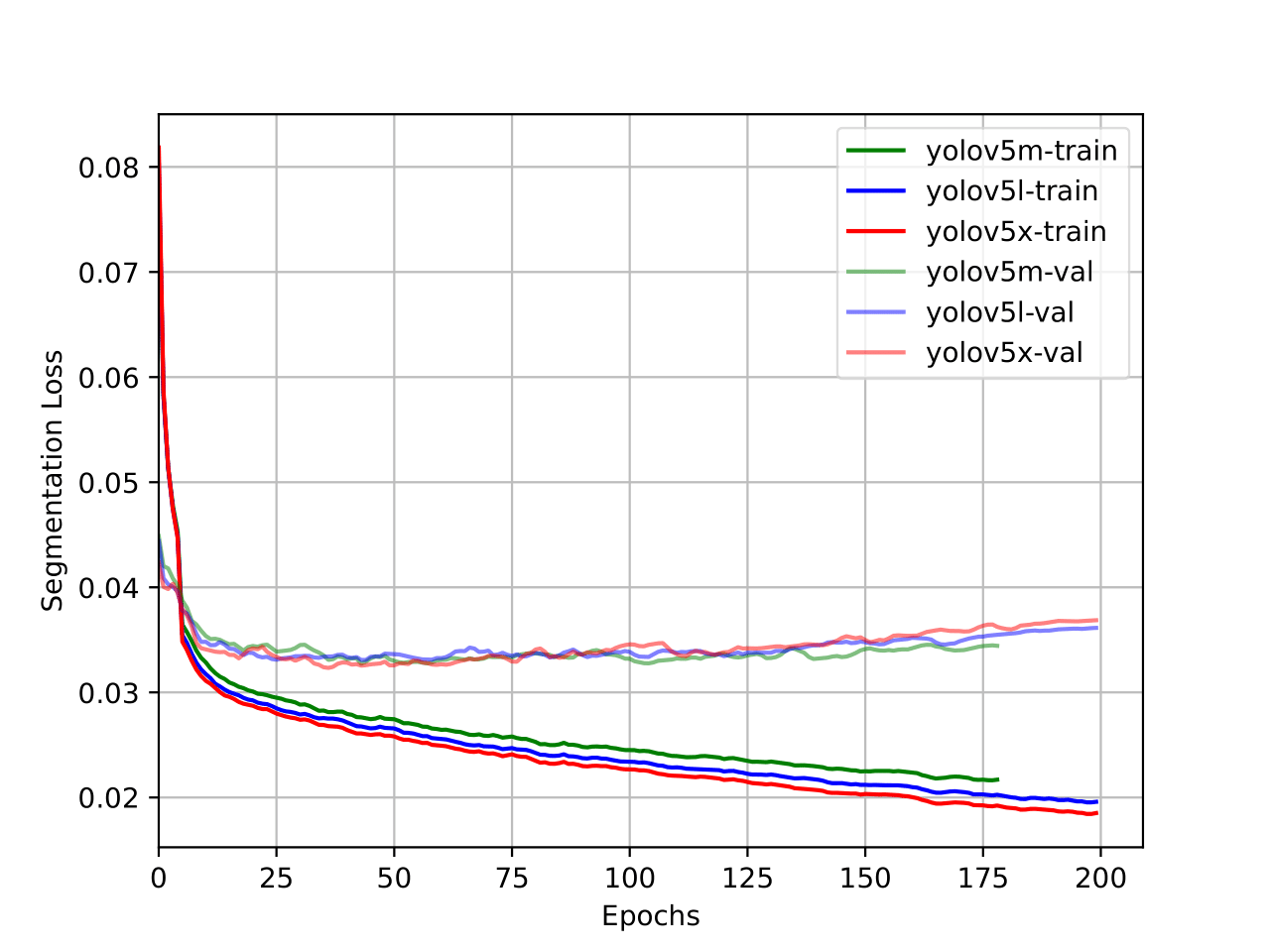}
    \caption{YOLOv5 segmentation loss decrease.}
    \label{fig:treining-6}
\end{subfigure}
\caption{Performance metrics for the YOLOv8 and YOLOv5 models during training and validation.}
\label{fig:allresults}
\end{figure*}

The neural networks trained on the temporal dataset demonstrated significant precision, recall, and mean average pixel precision throughout all stages of the plantation. This section provides an overview of the network training process and presents the achieved metrics.

The neural networks underwent training for 200 epochs, employing early stopping and a batch size of 8. The early stopping mechanism, with a patience hyperparameter set to 50 epochs, allowed for the termination of training when performance improvements became marginal. Pretraining the model on ImageNet~\cite{ridnik2021imagenet} provided a solid foundation for feature extraction. The training process utilized a constant learning rate of 0.01, an Adam optimizer with a momentum of 0.937, and a weight decay of 0.0005. Notably, dropout was not applied, and the training mode operated in a deterministic manner, ensuring reproducibility and stability throughout the training process. This comprehensive configuration aimed to strike a balance between model complexity, convergence speed, and generalization performance.

The learning curves can be seen in Fig.~\ref{fig:allresults}. In Figures \ref{fig:treining-2} and \ref{fig:treining-5}, the class loss curves for both the validation and training sets across all models are displayed. In Figures \ref{fig:treining-3} and \ref{fig:treining-6}, the loss curve for segmentation can be seen. The loss curve for class is measured using Distribution Focal Loss (the focal loss helps to address class imbalance in object detection by assigning different weights to easy and hard examples), and for segmentation is Binary Cross Entropy with Logits (binary classification loss function that combines sigmoid transformation of logits with binary cross-entropy calculation, it's used for stable and efficient training in binary classification tasks). These curves display a gradual and consistent decline, highlighting the models' efficacy in comprehending the data throughout the training process. Additionally, the mean average precision on both the training and validation sets is presented in Figures \ref{fig:treining-1} and \ref{fig:treining-4}, showcasing a steady and continuous upward trend, ultimately converging to yield strong results. The mAp-50 is calculated by averaging the precision values at the top 50 ranked predictions for a given task, providing a powerful performance metric for object detection.

\begin{table}[t!] % Use table for a single column
    \centering
    \setlength{\tabcolsep}{10pt}
    \caption{Box mAP50 for all models and classes.}
    \begin{tabular}{lccc}
        \toprule
        \textbf{Architecture} & \textbf{Caruru Weed} & \textbf{Grassy Weed} & \textbf{Soy Plant}\\
        \midrule
        YOLOv8m & \textbf{0.789} & \textbf{0.628} & \textbf{0.887} \\
        YOLOv8l & 0.759 & 0.593 & 0.880 \\
        YOLOv8x & 0.758 & 0.613 & 0.870 \\
        YOLOv5m & 0.744 & 0.623 & 0.866 \\
        YOLOv5l & 0.772 & 0.607 & 0.856 \\
        YOLOv5x & 0.720 & 0.606 & 0.860 \\
        \bottomrule
    \end{tabular}
    \label{tab:map50-values-box}
\end{table}

In Table~\ref{tab:map50-values-box}, mAP-50 metrics for bounding boxes in the test set are shown. The YOLOv8m model achieved the highest scores, with 0.789 for caruru weed objects, 0.628 for grassy weed objects, and 0.887 for Soy objects. Notably, the YOLOv8 models outperformed the YOLOv5 models across all classes, indicating a significant improvement.

Table~\ref{tab:precision-recall-box} presents the average bounding box precision and recall metrics for all classes across all models. YOLOv5m exhibited the highest average precision at 0.773, while YOLOv5l achieved the best recall at 0.738. Notably, YOLOv8 and YOLOv5 models demonstrated similar results in these metrics.

\begin{table}[t!] % Use table for a single column
    \centering
    \setlength{\tabcolsep}{12pt}
    \caption{Average box precision and recall for all models and classes.}
    \begin{tabular}{lcc}
        \toprule
        \textbf{Architecture} & \textbf{Precision (all classes)} & \textbf{Recall (all classes)}\\
        \midrule
        YOLOv8m & 0.767 & 0.737 \\
        YOLOv8l & 0.729 & 0.719 \\
        YOLOv8x & 0.772 & 0.693 \\
        YOLOv5m & \textbf{0.773} & 0.728 \\
        YOLOv5l & 0.770 & \textbf{0.738} \\
        YOLOv5x & 0.746 & 0.730 \\
        \bottomrule
    \end{tabular}
    \label{tab:precision-recall-box}
\end{table}

The segmentation mAP-50 values can be seen in Table~\ref{tab:map50-values-seg}, with YOLOv8m consistently outperforming other models across all classes. Specifically, YOLOv8m achieved a mAP of 0.787 for caruru and 0.696 for grassy weed, and an impressive 0.901 for Soy. Once more, the YOLOv8 models demonstrated superior performance compared to the YOLOv5 models.

\begin{table}[b!] % Use table for a single column
    \centering
    \setlength{\tabcolsep}{10pt}
    \caption{Segmentation mAP50 for all models and classes.}
    \begin{tabular}{lccc}
        \toprule
        \textbf{Architecture} & \textbf{Caruru Weed} & \textbf{Grassy Weed} & \textbf{Soy Plant}\\
        \midrule
        YOLOv8m & \textbf{0.787} & \textbf{0.696} & \textbf{0.901} \\
        YOLOv8l & 0.777 & 0.614 & 0.895 \\
        YOLOv8x & 0.761 & 0.591 & 0.888 \\
        YOLOv5m & 0.727 & 0.634 & 0.884 \\
        YOLOv5l & 0.742 & 0.610 & 0.875 \\
        YOLOv5x & 0.707 & 0.603 & 0.870 \\
        \bottomrule
    \end{tabular}
    \label{tab:map50-values-seg}
\end{table}

Precision and recall values for segmentation are provided at Table~\ref{tab:precision-recall-seg}. YOLOv8x achieved the highest precision at 0.791, while YOLOv8m attained the best recall at 0.733. Notably, precision and recall values across all models were quite similar for this task.

\begin{figure*}[tp]
    \centering
    \includegraphics[width=\linewidth]{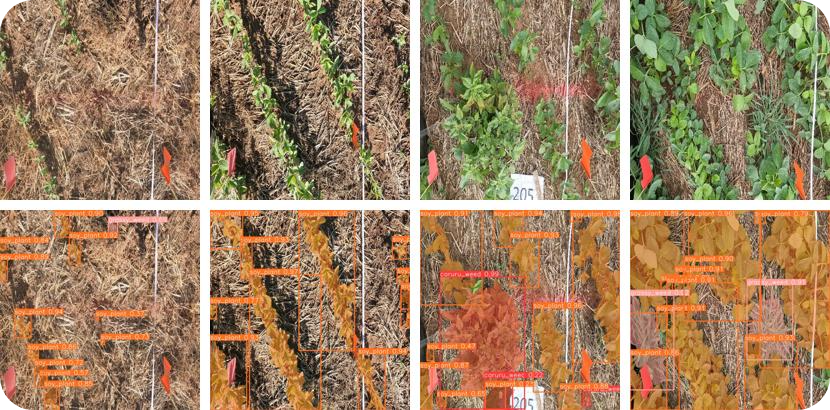}
    \caption{YOLOv8X model predictions over multiple stages of soy growth.}
    \label{fig:pred-sample}
\end{figure*}

To provide a more comprehensive illustration of model predictions on test images from the dataset, Fig.~\ref{fig:pred-sample} presents sample predictions for YOLOv8x across the initial, medium, and advanced stages of soy growth. This includes precise segmentation and classification of soy, caruru, and grassy weeds, demonstrating the model's accuracy in delineating these plants.

\section{Analysis of performance on initial and late stages}\label{sec:analysis}

In this section, the primary focus revolves around the neural network's role in effectively detecting weed presence. Our analysis will be dedicated to evaluating the performance of the YOLOv8 models on images featuring a coexistence of both weed and soy elements, through box predictions rather then mask predictions. We curated two distinct sets of 25 images, each depicting varying degrees of weed presence. The initial group comprises images captured during the seedling and early stages of plantation, while the second group encapsulates images from the harvest stages. Our objective is to test the YOLOv8 models' efficacy in accurately identifying weed instances in these diverse agricultural scenarios, and if there is a several difference in performance from early to late stages.

\begin{table}[b!] % Use table for a single column
    \centering
    \setlength{\tabcolsep}{12pt}
    \caption{Average segmentation precision and recall for all models and classes.}
    \begin{tabular}{lcc}
        \toprule
        \textbf{Architecture} & \textbf{Precision (all classes)} & \textbf{Recall (all classes)} \\
        \midrule
        YOLOv8m & 0.784 & \textbf{0.733} \\
        YOLOv8l & 0.743 & 0.732 \\
        YOLOv8x & \textbf{0.791} & 0.692 \\
        YOLOv5m & 0.770 & 0.721 \\
        YOLOv5l & 0.766 & 0.731 \\
        YOLOv5x & 0.743 & 0.716 \\
        \bottomrule
    \end{tabular}
    \label{tab:precision-recall-seg}
\end{table}

Considering the box mAP50 results in Table~\ref{tab:early-map50-values-box} and Table~\ref{tab:late-map50-values-box}, it is evident that the models perform better when both soy and weeds are present in the image. This improved performance might be attributed to the networks producing false positives in the absence of weeds. However, this is not a significant concern given the task at hand, especially since precision remains high in scenarios where only soy is present.

\begin{table}[t!] % Use table for a single column
    \centering
    \setlength{\tabcolsep}{10pt}
    \caption{Box mAP50 for all models and classes in early stages.}
    \begin{tabular}{lccc}
        \toprule
        \textbf{Architecture} & \textbf{Caruru Weed} & \textbf{Grassy Weed} & \textbf{Soy Plant}\\
        \midrule
        YOLOv8m & 0.995 & 0.915 & 0.936 \\
        YOLOv8l & 0.995 & 0.838 & 0.930 \\
        YOLOv8x & \textbf{0.995} & \textbf{0.981} & \textbf{0.946} \\
        \bottomrule
    \end{tabular}
    \label{tab:early-map50-values-box}
\end{table}

\begin{table}[b!] % Use table for a single column
    \centering
    \setlength{\tabcolsep}{10pt}
    \caption{Box mAP50 for all models and classes in late stages.}
    \begin{tabular}{lccc}
        \toprule
        \textbf{Architecture} & \textbf{Caruru Weed} & \textbf{Grassy Weed} & \textbf{Soy Plant}\\
        \midrule
        YOLOv8m & 0.970 & 0.884 & 0.905 \\
        YOLOv8l & 0.933 & 0.901 & 0.901 \\
        YOLOv8x & \textbf{0.984} & \textbf{0.914} & \textbf{0.958} \\
        \bottomrule
    \end{tabular}
    \label{tab:late-map50-values-box}
\end{table}

Another noteworthy observation from the tabulated data is the lack of significant disparities in model performance across different stages of plantation. This indicates that the dataset is diverse and that the trained models can effectively handle a variety of agricultural scenarios.

Considering the box precision and recall metrics in Table~\ref{tab:early-precision-recall-box} and Table~\ref{tab:late-precision-recall-box}, it is apparent that no single model distinctly outperforms the others. All YOLOv8 models consistently deliver reliable results in detecting both soy and weed in the images. A subtle trend shows a slight improvement in precision and recall metrics in the later stages of plantation, although this enhancement is not particularly significant.

\begin{table}[b!] % Use table for a single column
    \centering
    \setlength{\tabcolsep}{12pt}
    \caption{Average box precision and recall for all models and classes in early stages.}
    \begin{tabular}{lcc}
        \toprule
        \textbf{Architecture} & \textbf{Precision (all classes)} & \textbf{Recall (all classes)}\\
        \midrule
        YOLOv8m & 0.868 & \textbf{0.943} \\
        YOLOv8l & \textbf{0.954} & 0.873 \\
        YOLOv8x & 0.916 & 0.909 \\
        \bottomrule
    \end{tabular}
    \label{tab:early-precision-recall-box}
\end{table}

\begin{table}[h!] % Use table for a single column
    \centering
    \setlength{\tabcolsep}{12pt}
    \caption{Average box precision and recall for all models and classes in late stages.}
    \begin{tabular}{lcc}
        \toprule
        \textbf{Architecture} & \textbf{Precision (all classes)} & \textbf{Recall (all classes)}\\
        \midrule
        YOLOv8m & \textbf{0.885} & 0.841 \\
        YOLOv8l & 0.858 & 0.864 \\
        YOLOv8x & 0.879 & \textbf{0.933} \\
        \bottomrule
    \end{tabular}
    \label{tab:late-precision-recall-box}
\end{table}

In conclusion, this analysis highlights the consistent performance of the networks across different stages of plantation, demonstrating the dataset's diversity and the models' robust generalization capabilities. Additionally, it reveals a significant performance boost in images featuring both soy and weed, aligning with the primary objective of segmenting and detecting these elements. This reaffirms the models' effectiveness in achieving their intended purpose.

\section{Discussion}\label{discussion}

In this section, we discuss several key points that emerged from the results and analysis.

Firstly, it is evident that the YOLOv8 models consistently outperform the YOLOv5 models. However, it is intriguing that the YOLOv8m model, with significantly fewer parameters, can sometimes outperform the YOLOv8x model. Our best hypothesis is that the dataset, containing only three classes, may lead to overfitting in the more complex YOLOv8x model, resulting in slightly lower performance compared to the smaller YOLOv8m model in certain cases.

Another important observation is the distinct difference in both box and segmentation mAP50 metrics for different classes. The precision for soy is significantly higher than for the two weed classes. This discrepancy likely arises because the dataset was collected in a soy plantation, resulting in a higher prevalence of soy compared to weeds, both in terms of pixel count and presence percentage in images.

It is crucial to clarify that the primary goal of employing these models was not to achieve state-of-the-art results per se, but to validate our dataset. This is why we utilized existing models rather than developing a new one specifically tailored to this task.

\section{Conclusion}\label{sec:conclusion}

This paper introduces a dataset containing 1000 images capturing the entire life cycle of soy crops, from seedling to harvest, while also documenting weed infestation over time. Each image provides a comprehensive view of the plantation and includes instance segmentation labeling. This dataset holds promising scientific value for applications related to weed detection, soy monitoring, and serves as a robust benchmark for neural network architectures due to its comprehensiveness and practicality.

The paper further includes analysis and comparison of state-of-the-art YOLO models using this dataset, contrasting the latest model iteration with the previous YOLOv5. The models trained on this dataset demonstrate exceptional precision in plant segmentation, and validate the quality of the data.

Looking ahead, potential future work could explore the extension of this dataset's methodology for crop disease detection and yield prediction, capitalizing on the temporal aspect of the dataset, which offers valuable practical applications.

\section*{Acknowledgements}

This research was funded by: Topsector TKI AgroFood under grant agreement LWV20242 and Ministry of Agriculture, Nature and Food Quality grant agreement BO-69-001-005 for “Smart technology for soybean production”.

The authors express their gratitude for the support and data contributions from various organizations, including Wageningen University \& Research, the Aquarius Project at Universidade Federal de Santa Maria, the Advanced Farm 360 project, Rometron, and Stara. Special thanks are extended to Ard Nieuwenhuizen and Telmo Jorge Carneiro Amado for their invaluable support throughout the research process.

\bibliographystyle{IEEEtran}
\bibliography{sn-bibliography}

\begin{thebibliography}{10}
\providecommand{\url}[1]{#1}
\csname url@samestyle\endcsname
\providecommand{\newblock}{\relax}
\providecommand{\bibinfo}[2]{#2}
\providecommand{\BIBentrySTDinterwordspacing}{\spaceskip=0pt\relax}
\providecommand{\BIBentryALTinterwordstretchfactor}{4}
\providecommand{\BIBentryALTinterwordspacing}{\spaceskip=\fontdimen2\font plus
\BIBentryALTinterwordstretchfactor\fontdimen3\font minus \fontdimen4\font\relax}
\providecommand{\BIBforeignlanguage}[2]{{%
\expandafter\ifx\csname l@#1\endcsname\relax
\typeout{** WARNING: IEEEtran.bst: No hyphenation pattern has been}%
\typeout{** loaded for the language `#1'. Using the pattern for}%
\typeout{** the default language instead.}%
\else
\language=\csname l@#1\endcsname
\fi
#2}}
\providecommand{\BIBdecl}{\relax}
\BIBdecl

\bibitem{pagano2016importance}
M.~C. Pagano and M.~Miransari, ``The importance of soybean production worldwide,'' in \emph{Abiotic and biotic stresses in soybean production}.\hskip 1em plus 0.5em minus 0.4em\relax Elsevier, 2016, pp. 1--26.

\bibitem{kujawa2021artificial}
S.~Kujawa and G.~Niedba{\l}a, ``Artificial neural networks in agriculture,'' p. 497, 2021.

\bibitem{escamilla2020applications}
A.~Escamilla-Garc{\'\i}a, G.~M. Soto-Zaraz{\'u}a, M.~Toledano-Ayala, E.~Rivas-Araiza, and A.~Gast{\'e}lum-Barrios, ``Applications of artificial neural networks in greenhouse technology and overview for smart agriculture development,'' \emph{Applied Sciences}, vol.~10, no.~11, p. 3835, 2020.

\bibitem{williams2019robotic}
H.~A. Williams, M.~H. Jones, M.~Nejati, M.~J. Seabright, J.~Bell, N.~D. Penhall, J.~J. Barnett, M.~D. Duke, A.~J. Scarfe, H.~S. Ahn \emph{et~al.}, ``Robotic kiwifruit harvesting using machine vision, convolutional neural networks, and robotic arms,'' \emph{biosystems engineering}, vol. 181, pp. 140--156, 2019.

\bibitem{vulpi2021recurrent}
F.~Vulpi, A.~Milella, R.~Marani, and G.~Reina, ``Recurrent and convolutional neural networks for deep terrain classification by autonomous robots,'' \emph{Journal of Terramechanics}, vol.~96, pp. 119--131, 2021.

\bibitem{rawat2017deep}
W.~Rawat and Z.~Wang, ``Deep convolutional neural networks for image classification: A comprehensive review,'' \emph{Neural computation}, vol.~29, no.~9, pp. 2352--2449, 2017.

\bibitem{kamilaris2018review}
A.~Kamilaris and F.~X. Prenafeta-Bold{\'u}, ``A review of the use of convolutional neural networks in agriculture,'' \emph{The Journal of Agricultural Science}, vol. 156, no.~3, pp. 312--322, 2018.

\bibitem{long2015fully}
J.~Long, E.~Shelhamer, and T.~Darrell, ``Fully convolutional networks for semantic segmentation,'' in \emph{Proceedings of the IEEE conference on computer vision and pattern recognition}, 2015, pp. 3431--3440.

\bibitem{guo2018review}
Y.~Guo, Y.~Liu, T.~Georgiou, and M.~S. Lew, ``A review of semantic segmentation using deep neural networks,'' \emph{International journal of multimedia information retrieval}, vol.~7, pp. 87--93, 2018.

\bibitem{olsen2019deepweeds}
A.~Olsen, D.~A. Konovalov, B.~Philippa, P.~Ridd, J.~C. Wood, J.~Johns, W.~Banks, B.~Girgenti, O.~Kenny, J.~Whinney \emph{et~al.}, ``Deepweeds: A multiclass weed species image dataset for deep learning,'' \emph{Scientific reports}, vol.~9, no.~1, p. 2058, 2019.

\bibitem{wang2022weed25}
P.~Wang, Y.~Tang, F.~Luo, L.~Wang, C.~Li, Q.~Niu, and H.~Li, ``Weed25: A deep learning dataset for weed identification,'' \emph{Frontiers in Plant Science}, vol.~13, p. 1053329, 2022.

\bibitem{MIGNONI2022107756}
\BIBentryALTinterwordspacing
M.~E. Mignoni, A.~Honorato, R.~Kunst, R.~Righi, and A.~Massuquetti, ``Soybean images dataset for caterpillar and diabrotica speciosa pest detection and classification,'' \emph{Data in Brief}, vol.~40, p. 107756, 2022. [Online]. Available: \url{https://www.sciencedirect.com/science/article/pii/S2352340921010313}
\BIBentrySTDinterwordspacing

\bibitem{francis2019disease}
M.~Francis and C.~Deisy, ``Disease detection and classification in agricultural plants using convolutional neural networks—a visual understanding,'' in \emph{2019 6th international conference on signal processing and integrated networks (SPIN)}.\hskip 1em plus 0.5em minus 0.4em\relax IEEE, 2019, pp. 1063--1068.

\bibitem{bedi2021plant}
P.~Bedi and P.~Gole, ``Plant disease detection using hybrid model based on convolutional autoencoder and convolutional neural network,'' \emph{Artificial Intelligence in Agriculture}, vol.~5, pp. 90--101, 2021.

\bibitem{hasan2021survey}
A.~M. Hasan, F.~Sohel, D.~Diepeveen, H.~Laga, and M.~G. Jones, ``A survey of deep learning techniques for weed detection from images,'' \emph{Computers and Electronics in Agriculture}, vol. 184, p. 106067, 2021.

\bibitem{ramirez2020deep}
W.~Ramirez, P.~Achanccaray, L.~Mendoza, and M.~Pacheco, ``Deep convolutional neural networks for weed detection in agricultural crops using optical aerial images,'' in \emph{2020 IEEE Latin American GRSS \& ISPRS Remote Sensing Conference (LAGIRS)}.\hskip 1em plus 0.5em minus 0.4em\relax IEEE, 2020, pp. 133--137.

\bibitem{hu2021rice}
D.~Hu, C.~Ma, Z.~Tian, G.~Shen, and L.~Li, ``Rice weed detection method on yolov4 convolutional neural network,'' in \emph{2021 international conference on artificial intelligence, big data and algorithms (CAIBDA)}.\hskip 1em plus 0.5em minus 0.4em\relax IEEE, 2021, pp. 41--45.

\bibitem{sudars2020dataset}
K.~Sudars, J.~Jasko, I.~Namatevs, L.~Ozola, and N.~Badaukis, ``Dataset of annotated food crops and weed images for robotic computer vision control,'' \emph{Data in brief}, vol.~31, p. 105833, 2020.

\bibitem{alam2022tobset}
M.~S. Alam, M.~Alam, M.~Tufail, M.~U. Khan, A.~G{\"u}ne{\c{s}}, B.~Salah, F.~E. Nasir, W.~Saleem, and M.~T. Khan, ``Tobset: A new tobacco crop and weeds image dataset and its utilization for vision-based spraying by agricultural robots,'' \emph{Applied Sciences}, vol.~12, no.~3, p. 1308, 2022.

\bibitem{medhi2022psfd}
E.~Medhi and N.~Deb, ``Psfd-musa: A dataset of banana plant, stem, fruit, leaf, and disease,'' \emph{Data in Brief}, vol.~43, p. 108427, 2022.

\bibitem{moupojou2023fieldplant}
E.~Moupojou, A.~Tagne, F.~Retraint, A.~Tadonkemwa, D.~Wilfried, H.~Tapamo, and M.~Nkenlifack, ``Fieldplant: A dataset of field plant images for plant disease detection and classification with deep learning,'' \emph{IEEE Access}, vol.~11, pp. 35\,398--35\,410, 2023.

\bibitem{yang2023semantic}
J.~Yang, S.~Guo, M.~J. Bocus, Q.~Chen, and R.~Fan, ``Semantic segmentation for autonomous driving,'' in \emph{Autonomous Driving Perception: Fundamentals and Applications}.\hskip 1em plus 0.5em minus 0.4em\relax Springer, 2023, pp. 101--137.

\bibitem{zhou2023hybrid}
Z.~Zhou, J.~Zhang, and C.~Gong, ``Hybrid semantic segmentation for tunnel lining cracks based on swin transformer and convolutional neural network,'' \emph{Computer-Aided Civil and Infrastructure Engineering}, 2023.

\bibitem{anand2021agrisegnet}
T.~Anand, S.~Sinha, M.~Mandal, V.~Chamola, and F.~R. Yu, ``Agrisegnet: Deep aerial semantic segmentation framework for iot-assisted precision agriculture,'' \emph{IEEE Sensors Journal}, vol.~21, no.~16, pp. 17\,581--17\,590, 2021.

\bibitem{bosilj2020transfer}
P.~Bosilj, E.~Aptoula, T.~Duckett, and G.~Cielniak, ``Transfer learning between crop types for semantic segmentation of crops versus weeds in precision agriculture,'' \emph{Journal of Field Robotics}, vol.~37, no.~1, pp. 7--19, 2020.

\bibitem{roboflow2023}
Roboflow, ``Roboflow: A tool for computer vision workflow,'' 2023, https://roboflow.com/.

\bibitem{lin2022roboflow}
Q.~Lin, G.~Ye, J.~Wang, and H.~Liu, ``Roboflow: a data-centric workflow management system for developing ai-enhanced robots,'' in \emph{Conference on Robot Learning}.\hskip 1em plus 0.5em minus 0.4em\relax PMLR, 2022, pp. 1789--1794.

\bibitem{yolov52020}
G.~Jocher \emph{et~al.}, ``Yolov5: Fifth version of you only look once for real-time object detection,'' 2020, https://github.com/ultralytics/yolov5.

\bibitem{ultralytics2023}
Ultralytics, ``Ultralytics - yolov8,'' 2023, https://github.com/ultralytics/ultralytics.

\bibitem{ridnik2021imagenet}
T.~Ridnik, E.~Ben-Baruch, A.~Noy, and L.~Zelnik-Manor, ``Imagenet-21k pretraining for the masses,'' \emph{arXiv preprint arXiv:2104.10972}, 2021.

\end{thebibliography}

\end{document}